\documentclass[runningheads]{llncs}
\usepackage[T1]{fontenc}
\usepackage{epsfig} 
\usepackage{amsmath,amsfonts,amssymb,mathtools}
\usepackage{algorithm,algorithmicx,algpseudocode}
\usepackage[utf8]{inputenc} 
\usepackage[T1]{fontenc}
\usepackage{url}
\usepackage{tabularx}
\usepackage{booktabs}
\usepackage{xcolor}
\usepackage{hyperref}
\usepackage[misc,geometry]{ifsym}
\usepackage{color}

\usepackage{graphicx,float}
\graphicspath{{figures/}}

\usepackage{subcaption}
\usepackage{float}

\usepackage{tikz}
\usetikzlibrary{backgrounds,angles,quotes,calc,patterns,decorations.pathmorphing,decorations.markings,positioning}

\usepackage{todonotes}

\def\cN{\mathcal{N}}
\def\cF{\mathcal{F}}

\def\cJ{\mathcal{J}}
\def\-{\text{-}}
\def\+{\text{+}}
\def\tm{\! - \!}
\def\tp{\! + \!}

\newcommand\given[1][]{\:#1\vert\:}

\begin{document}
\title{Coupled autoregressive active inference agents\\ for control of multi-joint dynamical systems}
\titlerunning{Coupling autoregressive active inference agents}

\author{Tim N. Nisslbeck\inst{1}\orcidID{0009-0007-3114-812X} \and \\
Wouter M. Kouw\inst{1}\orcidID{0000-0002-0547-4817}}
\authorrunning{T.N. Nisslbeck \& W.M. Kouw}
%
\institute{Bayesian Intelligent Autonomous Systems lab, TU Eindhoven, Netherlands 
\email{t.n.nisslbeck@tue.nl}\textsuperscript{(\Letter)}, \email{w.m.kouw@tue.nl}}
\maketitle              
\begin{abstract}
We propose an active inference agent to identify and control a mechanical system with multiple bodies connected by joints. This agent is constructed from multiple scalar autoregressive model-based agents, coupled together by virtue of sharing memories. Each subagent infers parameters through Bayesian filtering and controls by minimizing expected free energy over a finite time horizon. We demonstrate that a coupled agent of this kind is able to learn the dynamics of a double mass-spring-damper system, and drive it to a desired position through a balance of explorative and exploitative actions. 
It outperforms the uncoupled subagents in terms of surprise and goal alignment.
\end{abstract}
\section{Introduction}
Our society relies heavily on mechatronic systems for manufacturing, energy, transport, logistics and healthcare. These systems are still largely designed using physics-driven models, offline system identification and optimal control techniques. However, this design framework leads to systems that tend to be sensitive to "noise", i.e., sensor and actuator imperfections, external disturbances, and unmodeled physics (e.g., heat, vibrations). Robustness requires adaptation to a changing environment by updating a model rapidly, continuously and data-efficiently. This is exactly what embodied artificial intelligence and cognitive robotics strive to achieve \cite{liagkou2021challenges,krichmar2018neurorobotics}. 
Reinforcement learning is a prime candidate framework, but it tends to be costly in terms of computational resources and training time \cite{bucak2001reinforcement}. A more appropriate framework for resource-constrained mechatronic systems is active inference, which characterizes itself by including optimal information gain in its data acquisition protocol \cite{parr2022active,parr2024active}.
Here we present scalar active inference agents that are coupled together to jointly control a mechatronic system with multiple inputs and multiple outputs \cite{massioni2009distributed}.

Active inference draws its roots from cognitive science where it is a process theory for intelligent behaviour \cite{friston2010action}. Many agents with discrete state and action spaces have been proposed as models of learning, exploration and curiosity \cite{friston2015active,friston2016active,schwartenbeck2019computational,da2020active}. The engineering community wants to use active inference as a framework for designing intelligent autonomous systems with continuous state and action spaces \cite{pio2016active,lanillos2021active,baioumy2021active,baltieri2019pid,imohiosen2020active}. A major challenge in designing such agents are the calculations of the differential entropies involved. Many models assume some form of Gaussian state transition or likelihood, often with parameters shaped by neural networks \cite{ueltzhoffer2018deep,van2020deep,huebotter2023learning}.
We build on recent work using autoregressive models fit for resource-constrained mechatronic systems \cite{kouw2023information}.
Our contributions include:
\begin{itemize}
    \item The formulation of a coupled active inference agent consisting of two scalar agents that share memories (Sec.~\ref{sec:coupling}).
    \item An empirical evaluation of coupled versus uncoupled agents on a double mass-spring-damper system (Sec.~\ref{sec:experiments}).
\end{itemize}

\section{Problem statement}
\begin{figure}[htb]
    \centering
     \begin{tikzpicture}[
 squarednode/.style={rectangle, draw=black, thick, minimum
 width=25mm,minimum height=8mm},
]

 \tikzstyle{spring}=[thick,decorate,decoration={zigzag,pre length=0.3cm,post
 length=0.3cm,segment length=6}]

 \tikzstyle{damper}=[thick,decoration={markings,  
   mark connection node=dmp,
   mark=at position 0.5 with 
   {
     \node (dmp) [thick,inner sep=0pt,transform shape,rotate=-90,minimum
 width=15pt,minimum height=3pt,draw=none] {};
     \draw [thick] ($(dmp.north east)+(2pt,0)$) -- (dmp.south east) -- (dmp.south
 west) -- ($(dmp.north west)+(2pt,0)$);
     \draw [thick] ($(dmp.north)+(0,-5pt)$) -- ($(dmp.north)+(0,5pt)$);
   }
 }, decorate]

 \tikzstyle{ground}=[fill,pattern=north east lines,thick,minimum width=0.75cm,minimum height=0.3cm]

 \node (M0) {};
 \node[squarednode] (M1) [below = 12mm of M0] {$m_1$};
 \node[squarednode] (M2) [below = 12mm of M1] {$m_2$};

 \draw[ground] ($(M0) - (2,0)$) -- ($(M0) + (2,0)$);
 \draw[spring] ($(M0.south) - (0.5,-.1)$) -- ($(M1.north) - (0.5,0)$) node [midway,left = 2mm] {$k_1$};
 \draw[spring] ($(M1.south) - (0.5,0)$) -- ($(M2.north) - (0.5,0)$) node [midway,left = 2mm] {$k_2$};
 
 \draw[damper] ($(M0.south) + (0.5,0.1)$) -- ($(M1.north) + (0.5,0)$) node [midway,right = 3mm] {$c_1$};
 \draw[damper] ($(M1.south) + (0.5,0)$) -- ($(M2.north) + (0.5,0)$) node [midway,right = 3mm] {$c_2$};
\draw[-latex,very thick] (2,-2) -- (2,-4) node[midway,right]{$g$};
 

 \end{tikzpicture}
    \quad
    \begin{tikzpicture}[thick]

\newcommand{\angA}{30}
\newcommand{\angB}{30}

\draw (-1,0) -- (1,0);

\begin{scope}[rotate=\angA]

\draw[fill=gray] (-0.2,0) rectangle (0.2,-2.5) 
	node[midway](a){$\bullet$};

\draw[fill=gray,
	rotate around={\angB:(0,-2.5)}
] (-0.2,-2.5) rectangle (0.2,-5)
	node[midway](b){$\bullet$};

\draw[fill=white] (0,-2.5) circle(0.2);
\draw[fill=white] (0,0) circle(0.2);
\draw[fill=white] (1.25,-4.65) circle(0.2);

\end{scope}

\draw[-latex,very thick] (a.center) -- ++(0,-1.25)
	node[midway,left]{$m_1 g$};
\draw[-latex,very thick] (b.center) -- ++(0,-1.25) 
	node[midway,left]{$m_2 g$};

 \draw[-latex,very thick, rotate around={\angA:(0.4,0)}] (0.4,0.25) -- (0.4,-2.25)
	node[midway,right]{$l_1$};
 \draw[-latex,very thick, rotate around={\angA+\angB:(1.6,-1.9)}] (1.6,-1.9) -- (1.6,-4.1)
	node[midway,right = 2mm]{$l_2$};

\end{tikzpicture}
    \caption{(Left) A double mass-spring-damper system where block $1$ is attached to a stationary frame and block $2$ is attached to the first block. The dynamics of the system are determined by the masses $m_i$ of the blocks, the stiffness of the springs $k_i$, the amount of friction $c_i$ the dampeners provide and gravity $\mathrm{g}$. (Right) A double compound pendulum system consisting of two single compound pendulums joined end-to-end. The dynamics of the system are determined by the masses $m_i$ and lengths $l_i$ of the poles.}
    \label{fig:system}
\end{figure}
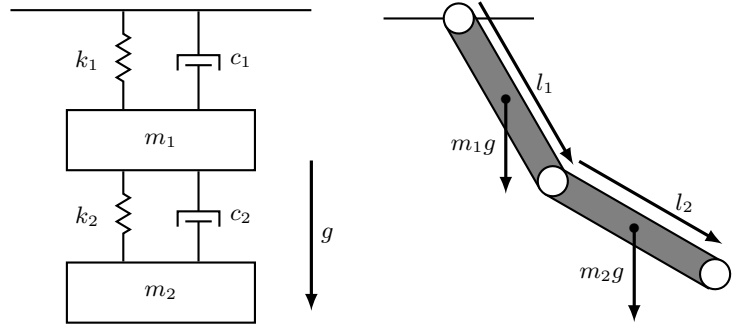

We study the class of multi-joint dynamical systems, characterized by simple mechanical systems connected in sequence. For example, a double mass-spring-damper system consists of one mass attached to a base through a spring and an accompanying damper, with a second mass connected to the first mass through another spring and damper (Figure \ref{fig:system} left). Similarly, a double pendulum consists of a single pendulum attached to a base and another single pendulum attached to the end of the first pendulum (Figure \ref{fig:system} right). The task is to find control policies for each motor such that the multi-joint dynamical system moves to a desired position. 
We expect that coupling agents together lets them more accurately predict joint motion and infer an appropriate control policy sooner.

\section{Agent specification}
Consider an agent, operating in discrete time, that sends inputs $u_k \in \mathbb{R}$ (a.k.a. controls, actions) to a system and measures its output $y_k \in \mathbb{R}$. The agent must drive the system to a desired output $y_{*}$ without knowledge of its dynamics.
Since this active inference agent minimizes expected free energy (EFE) based on an autoregressive exogenous (ARX) model, we refer to it as an ARX-EFE agent.

\subsection{Probabilistic model}
We specify a likelihood function of the form:
\begin{align} \label{eq:likelihood}
    p(y_{k} \given \theta, \tau, u_{k}, \bar{u}_k, \bar{y}_{k}) &= \mathcal{N}\big(y_{k} \given \theta^{\intercal} \begin{bmatrix} u_{k} \ \bar{u}_{k} \ \bar{y}_k \end{bmatrix}, \tau^{-1} \big) \, ,
\end{align}
where the vectors $\bar{y}_k \in \mathbb{R}^{M_y}$ and $\bar{u}_k \in \mathbb{R}^{M_u}$ are buffers containing previous observations of the system outputs and inputs, where $M_y$ and $M_u$ are the lengths of the output and input buffers, respectively.
This defines the above likelihood as an autoregressive model. $\theta \in \mathbb{R}^{D}$, where $D = M_y + M_u + 1$, are coefficients and $\tau \in \mathbb{R}^{+}$ represents a precision parameter.

The prior distribution on the parameters is a multivariate Gaussian - univariate Gamma distribution \cite[ID: D5]{Soch}:
\begin{align} \label{eq:priors}
    p(\theta, \tau) \triangleq \mathcal{NG}\big(\theta, \tau \given \mu_0, \Lambda_0, \alpha_0, \beta_0 \big) = \mathcal{N}(\theta \given \mu_0, (\tau \Lambda_0)^{-1} \big) \, \mathcal{G}\big(\tau \given \alpha_0, \beta_0 \big) \, .
\end{align}
The prior distributions over inputs are assumed to be independent over time:
\begin{align}
    p(u_k) \triangleq \cN(u_k \given 0, \eta^{-1}) \label{eq:control-prior} \, ,
\end{align}
with precision parameter $\eta$. This choice has a regularizing effect on the inferred controls (Sec.~\ref{sec:fe-controls}). 

\subsection{Inference} 
Our inference procedure is separated into a parameter belief update procedure given observed data, and control estimation given parameters.

\subsubsection{Parameters} \label{sec:fe-params}
First, note that, at time $k$, the control $u_k$ has been executed and is known to the agent. Henceforth, we shall use $\hat{u}_k$ and $\hat{y}_k$ to differentiate observed variables from unobserved ones. Furthermore, let
\begin{align}
    x_k = \begin{bmatrix} u_k \ \bar{u}_k \ \bar{y}_k \end{bmatrix} \, .
\end{align}
The parameter posterior distribution is obtained by Bayesian filtering \cite{sarkka2013bayesian}:
\begin{align} \label{eq:bayesian-filtering}
    \underbrace{p\big(\theta, \tau \given \mathcal{D}_{k} \big)}_{\text{posterior}} \! = \! \frac{\overbrace{p\big(\hat{y}_k \given \theta, \tau, \hat{u}_k, \bar{u}_{k}, \bar{y}_k \big)}^{\text{likelihood}}}{\underbrace{p\big(\hat{y}_{k} \given \hat{u}_{k}, \mathcal{D}_{k\-1}\big)}_{\text{evidence}}} \underbrace{p\big(\theta, \tau \given \mathcal{D}_{k\-1}\big)}_{\text{prior}} . 
\end{align}
where $\mathcal{D}_k = \{\hat{y}_i, \hat{u}_i\}_{i=1}^{k}$ is the data up to time $k$. The evidence (a.k.a. marginal likelihood) is
\begin{align} \label{eq:evidence-filtering}
    p\big(\hat{y}_{k} \given \hat{u}_{k}, \mathcal{D}_{k\-1}\big)  =  \int p\big(\hat{y}_{k} \given \theta, \tau, \hat{u}_k, \bar{u}_{k}, \bar{y}_k \big) p\big(\theta,\tau \given \mathcal{D}_{k\-1}\big) \, \mathrm{d}(\theta,\tau) .
\end{align}
We obtain an exact posterior distribution using the multivariate Gaussian - univariate Gamma prior distribution specified in Eq.~\ref{eq:priors} \cite{kouw2023information}:
\begin{align}
    p(\theta, \tau \given \mathcal{D}_k) = \mathcal{NG}(\theta, \tau \given \mu_k, \Lambda_k, \alpha_k, \beta_k) \, .
\end{align}
where
\begin{align} \label{eq:truepost_params}
    &\mu_k = \big(x_k x_k^{\intercal} \! + \! \Lambda_{k\-1} \big)^{-1}\big(x_k \hat{y}_k \! + \! \Lambda_{k\-1} \mu_{k\-1} \big) \, , \quad
    \Lambda_k = x_k x_k^{\intercal}  +  \Lambda_{k\-1} , \\ 
    &\alpha_k = \alpha_{k\-1} + \frac{1}{2} \, , \quad
    \beta_k = \beta_{k\-1} +  \frac{1}{2}\big( \hat{y}_k^2  -  \mu_k^{\intercal} \Lambda_k \mu_k + \mu_{k\-1}^{\intercal}\Lambda_{k\-1} \mu_{k\-1} \big) \, .
\end{align}
The marginal posterior distributions are Gamma distributed and multivariate location-scale T-distributed \cite[ID: P36]{Soch}:
\begin{align}
    p(\tau \given \mathcal{D}_k)  &= \int  p(\theta, \tau \given \mathcal{D}_k) \mathrm{d}\theta = \mathcal{G}(\tau \given \alpha_k, \beta_k) \label{eq:margpost-tau} \, , \\
    p(\theta \given \mathcal{D}_k)  &=  \int p(\theta, \tau \given \mathcal{D}_k) \mathrm{d}\tau = \mathcal{T}_{2\alpha_k}\big(\theta \given \mu_k, \frac{\beta_k}{\alpha_k} \Lambda_k^{\-1}\big) . \label{eq:margpost-theta}
\end{align}
The $2\alpha_k$ subscript refers to the T-distribution's degrees of freedom parameter.

\subsubsection{Controls} \label{sec:fe-controls}
In order to effectively drive the system to the goal, the agent must make accurate predictions for future outputs. The predictive probability of the input, output and parameters at time $t = k+1$ is:
\begin{align} \label{eq:p-future}
    p(y_t, \theta, \tau, u_t \given \mathcal{D}_k) &= p(y_t \given \theta, \tau, u_t, \bar{u}_{t}, \bar{y}_t) \,  p(\theta, \tau \given \mathcal{D}_k) p(u_t) \, .
\end{align}
Note that at time $t=k+1$, the buffers $\bar{y}_t = [\hat{y}_{k} \ \hat{y}_{k-1} \ \dots]$ and $\bar{u}_t = [\hat{u}_{k} \ \hat{u}_{k-1} \ \dots]$ contain only observed variables (i.e., there are no products between random variables).
To incorporate the goal output, we invert (see Eq.~\ref{eq:futureparampost-bayes}) the conditional dependency in the predictive probability for the output and parameters:
\begin{align} \label{eq:decomp-futurejoint}
    p(y_t \given \theta, \tau, u_t, \bar{u}_{t}, \bar{y}_t) \,  p(\theta, \tau \given \mathcal{D}_k)
    &= p(y_t, \theta, \tau \given u_t; \mathcal{D}_k) \\
    &= p(\theta, \tau \given y_t, u_t; \mathcal{D}_k) \, p(y_t) \, .
\end{align}
We intervene on the marginal prior distribution over future output, $p(y_t)$, with our chosen goal prior parameters:
\begin{align} 
    p(y_t) \rightarrow p(y_t \given y_{*}) \triangleq \cN(y_t \given m_{*}, v_{*}) \, . \label{eq:goalp}
\end{align}
Now, to infer a posterior distribution for the control variable $u_t$, we introduce an expected free energy functional \cite{friston2015active,van2022active},
\begin{align} \label{eq:EFE0}
    \mathcal{F}_k[q]  \triangleq   \mathbb{E}_{q(y_t, \theta, \tau, u_t)} \Big[ \ln \frac{p(\theta, \tau \given \mathcal{D}_k)q(u_t)}{p(\theta, \tau \given y_t, u_t; \mathcal{D}_k) p(y_t | y_{*}) p(u_t)} \Big] ,
\end{align}
with a variational model of the form:
\begin{align} \label{eq:q-factorization}
    q(y_t, \theta, \tau, u_t) \triangleq p(y_t, \theta, \tau \given u_t; \mathcal{D}_k) q(u_t) \, .
\end{align}
Inferring the optimal control at time $t$ refers to minimizing the free energy functional with respect to the variational distribution $q(u_t)$:
\begin{align}
    q^{*}(u_t) = \underset{q \, \in \, Q}{\arg \min} \ \mathcal{F}_k[q] \, .
\end{align}
where $Q$ represents the space of candidate distributions.
We can re-arrange the free energy functional to simplify the variational minimization problem:
\begin{align}
\mathbb{E}_{q(y_t, u_t, \theta, \tau)} &\Big[ \ln \frac{p(\theta, \tau \given \mathcal{D}_k) \, q(u_t)}{p(\theta, \tau \given y_t, u_t; \mathcal{D}_k) p(y_t \given y_{*}) p(u_t)} \Big] = \\
&  \mathbb{E}_{q(u_t)} \Big[ \underbrace{\mathbb{E}_{p(y_t, \theta, \tau \given u_t; \mathcal{D}_k)} \big[ \ln \frac{p(\theta, \tau \given \mathcal{D}_k)}{p(\theta, \tau \given y_t, u_t; \mathcal{D}_k) p(y_t | y_{*})} \big]}_{\cJ_k(u_t)} +  \ln \frac{q(u_t)}{p(u_t)} \Big] . \nonumber
\end{align}
Using $\cJ_k(u_t) = \ln (1/ \exp(-\cJ_k(u_t)))$, the expected free energy functional can be expressed as a Kullback-Leibler divergence
\begin{align} \label{eq:EFE2}
    \cF_k[q] = \mathbb{E}_{q(u_t)} \Big[ \ln \frac{q(u_t)}{\exp\big(-\cJ_k(u_t) \big) p(u_t)} \Big] \, ,
\end{align}
which is minimal when $q^{*}(u_t) =  \exp\big(- \cJ_k(u_t) \big) p(u_t)$ \cite{mackay2003information}. Thus, we have an optimal approximate posterior distribution over controls. 

The only unknown distribution in $\cJ_k(u_t)$ is the distribution over parameters given the future output and control (see Eq.~\ref{eq:decomp-futurejoint}). It can be related to known distributions through Bayes' rule:
\begin{align}
    p(\theta, \tau &\given y_t, u_t; \mathcal{D}_k) =  \frac{p(y_t \given \theta, \tau, u_t, \bar{u}_t, \bar{y}_t)\ p(\theta, \tau \given \mathcal{D}_k)}{\int p(y_t \given \theta, \tau, u_t, \bar{u}_t, \bar{y}_t) p(\theta,\tau \given \mathcal{D}_k) \mathrm{d}(\theta,\tau)} . \label{eq:futureparampost-bayes}
\end{align}
The distribution that results from the marginalization in the denominator is the posterior predictive distribution $p(y_t \given u_t; \mathcal{D}_k)$ and can be derived analytically within our model specification \cite{kouw2023information}:
\begin{align} 
    p(y_t \given u_t; \mathcal{D}_k) &\triangleq \int p(y_t \given \theta, \tau, u_t, \bar{u}_t, \bar{y}_t) p(\theta,\tau \given \mathcal{D}_k) \mathrm{d}(\theta,\tau)  \\
    &= \mathcal{T}_{2\alpha_k} \Big(y_t \given \mu_k^{\intercal} x_t, \, \frac{\beta_k}{\alpha_k}\big(x_t^{\intercal} \Lambda_k^{-1} x_t + 1\big) \Big) \, , \label{eq:q-postpred}
\end{align}
for $x_t = \begin{bmatrix} u_t \ \bar{u}_t \ \bar{y}_t \end{bmatrix}$.
If we replace $p(\theta, \tau \given y_t, u_t)$ in the expected free energy function with the right-hand side of Eq.~\ref{eq:futureparampost-bayes} and use Eq.~\ref{eq:p-future}, then it can be split into two components:
\begin{align}
\cJ_k(u_t) 
    &= \mathbb{E}_{p(y_t \given u_t; \mathcal{D}_k)} \Big[-\ln p(y_t \given y_{*}) \Big] \label{eq:EFE4} \\
    &\qquad - \mathbb{E}_{p(y_t, \theta, \tau \given u_t; \mathcal{D}_k)} \Big[\ln \frac{p(y_t, \theta, \tau \given u_t; \mathcal{D}_k) }{p(\theta, \tau \given \mathcal{D}_k) p(y_t \given u_t; \mathcal{D}_k)} \Big] \, . \nonumber
\end{align}
One may recognize the first term as a cross-entropy, describing the dissimilarity between the posterior predictive distribution and the goal prior distribution \cite{mackay2003information}. The second term is the mutual information between the parameter posterior and the predictive distribution, describing how much information is gained on the parameters upon measuring a system output \cite{mackay2003information}.
Solving the expectations yields 
\begin{align} \label{eq:EFE-final}
    \cJ_k(u_t) \! = \!  \mathrm{C} \! + \!  \frac{1}{2 v_{*}} \big((\mu_k^{\intercal} x_t \! - \! m_{*} )^2  \! + \!  \frac{\beta_k}{\alpha_k \- 1}(x_t^{\intercal} \Lambda_k^{-1} x_t \! + \! 1) \big) \! - \! \frac{1}{2} \ln (x_t^{\intercal}\Lambda_k^{-1}x_t \! + \! 1 ) ,
\end{align}
where $\mathrm{C}$ are constants that do not depend on $u_t$ \cite{kouw2023information}.

Unfortunately, the functional form of $q^{*}(u_t)$ does not appear to be a member of a known parametric family. This means we do not have access to analytic solutions of the moments of this distribution. If only its most probable value is of interest, then the most straightforward approach is maximum a posteriori (MAP) estimation. The MAP estimator can be written as a minimization over a negative logarithmic transformation of $q^{*}(u_t)$:
\begin{align} \label{eq:neglog}
    \hat{u}_t &= \underset{u_t \, \in \, \mathcal{U}}{\arg \max} \, q^{*}(u_t) \\
    &=  \underset{u_t \, \in \, \mathcal{U}}{\arg \min} \ \cJ_k(u_t) - \ln p(u_t) \, ,\label{eq:u_MAP}
\end{align}
where $\mathcal{U} = \{ u \in \mathbb{R} \given u_{min} \leq u \leq u_{max}\}$ refers to the space of affordable controls. It can be used to incorporate practical constraints such as torque limits. If an approximate uncertainty over the controls is required, then the above MAP estimate can be extended to a Laplace approximation \cite{friston2007variational}.

\subsection{Coupling} \label{sec:coupling}
The above ARX-EFE agent is scalar and can only operate on single-input single-output systems. 
We can of course naively group multiple such agents together to operate on a multi-input multi-output system, as is sometimes done with Gaussian processes \cite[Sec.~9.1]{williams2006gaussian}. But that ignores correlations between outputs which is important for prediction of motion in mechanical systems. We propose to couple agents together by virtue of incorporating additional signals into the autoregressive data buffers (i.e., memories) $x_t$. For agent $j \neq i$, sharing the output buffer between agents would take the form of:
\begin{align} \label{eq:coupled-likelihood}
    p(y_{i,k} \given \theta_i, \tau_i, u_{i,k}, \bar{u}_{i,k}, \bar{y}_{i,k}, \bar{y}_{j,k}) &= \mathcal{N}\big(y_{i,k} \given \theta_i^{\intercal} \begin{bmatrix} u_{i,k} \ \bar{u}_{i,k} \ \bar{y}_{i,k}, \ \bar{y}_{j,k} \end{bmatrix}, \tau^{-1} \big) \, .
\end{align}

Through sharing data buffers, the prediction for one system component will depend explicitly on another component.
However, this solution poses a problem for when the agent wants to extend its time horizon to $t > k+1$. In principle, due to the independence assumptions on the prior $p(u_t)$ and the variational control posteriors $q(u_t)$, the joint control posterior distribution can be formed as:
\begin{align}
    q^{*}(u_t, \dots, u_{t+T}) = \prod_{t=1}^T \, p(u_t) \, \exp\big(-\cJ_k(u_t) \big) \, .
\end{align}
For a single agent, the output buffer for $t > k+2$ can be filled with the maximum a posteriori value of its prediction at $t = k+1$ \cite{kouw2023information}. This solution can be applied recursively so the time horizon can be extended arbitrarily far. However, in a coupled setting, agent $1$ has to use agent $2$'s prediction for $k+1$. But agent $2$'s prediction depends on agent $1$'s action. Thus, the coupled agents must solve a \emph{nested} optimization procedure, iteratively alternating between two scalar optimization procedures. This means coupling becomes computationally expensive for time horizons $t > k+1$. 

\subsection{Optimization} \label{sec:optimization}
The optimization problem in Eq.~\ref{eq:u_MAP} can be solved in a number of ways. Firstly, using modern automatic or algorithmic differentiation tools, the gradient and Hessian with respect to $u_t$ can be obtained. Iterative procedures such as gradient descent or (quasi-)Newton methods, will then return approximate minimizers. The most straightforward way to enforce control space constraints is to utilize an interior-point method \cite{gill2019practical}. Such a method imposes a log-barrier function, which increases an objective function drastically as it approaches the constraint boundary. 

Alternatively, one could quantize the control space $\mathcal{U}$, calculate Eq.~\ref{eq:u_MAP} for every possible value and select the minimizer. For a single time-step and a scalar control, this procedure may actually be computationally cheaper as it does not require iteration. It does come at the cost of a quantization error for the estimated $\hat{u}_t$, and, of course, it does not scale well for longer time horizons due to the curse of dimensionality (the discretization interval becomes a tensor).

\section{Experiments} \label{sec:experiments}

\subsection{System description}
We perform an experiment\footnote{Code found at \href{https://github.com/biaslab/IWAI2024-CARXEFE}{https://github.com/biaslab/IWAI2024-CARXEFE}} on a double mass-spring-damper system.
Its equation of motions are the following second-order ordinary differential equations (ODE) \cite{lopes2010bayesian}:
\begin{align} \label{eq:eom}
    \begin{bmatrix}
    m_1 & 0 \\
    0 & m_2
    \end{bmatrix}
    \begin{bmatrix} \ddot{z}_1 \\ \ddot{z}_2 \end{bmatrix} =
    \begin{bmatrix}
    \tm (c_1 \tp c_2) &     c_2 \\
                 c_2  & \tm c_2
    \end{bmatrix}
    \begin{bmatrix} \dot{z}_1 \\ \dot{z}_2 \end{bmatrix}
    \tp
    \begin{bmatrix}
    \tm (k_1 \tp k_2) &     k_2 \\
                 k_2  & \tm k_2
    \end{bmatrix}
    \begin{bmatrix} z_1 \\ z_2 \end{bmatrix}
    \tp
    \begin{bmatrix} u_1 \\ u_2 \end{bmatrix},
\end{align}
where each block $i$ has displacement (or position) $z_i$, velocity $\dot{z_i}$, acceleration $\ddot{z_i}$, mass $i$, damping coefficient $c_i$, spring coefficient $k_i$, and external force (control) $u_i$.
In our experiments, we choose $c_1 = c_2 = 0.1$, $k_1 = k_2 = 1.0$, and $m_1 = m_2 = 1.0$.
To update the state of the system, we numerically solve the ODE using the second-order St\o rmer-Verlet integration method \cite{press2007numerical}.
This method involves updating the displacement of the mass as follows:
\begin{equation} \label{eq:eom-integration}
z_{t+1} = z_t + \Delta t \dot{z_t} + \frac{1}{2} \Delta t^2 \ddot{z_t},
\end{equation}
where $\ddot{z_t}$ is calculated from the equations of motion in Eq.~\ref{eq:eom}.
The initial state of the system is the fixed point $z_0 = [0.0, 0.0, 0.0, 0.0]$.
By reducing the step size $\Delta t$ and correspondingly increasing the number of updates $n_{iter}$, we can reduce the risk of numerical instabilities.
In our experiments, we choose $\Delta t = 0.01$ and $n_{iter} = 120$.
The observed measurement $y_t$ at time $t$ of the system is the position $z_t$ plus measurement noise $\varepsilon \sim \mathcal{N}(0, \sigma_\varepsilon^2 I_{M_y})$, where $I_{M_y}$ is an identity matrix of size $M_y \times M_y$, and $\sigma_\varepsilon^2$ is the variance of the noise.
We use $\sigma_\varepsilon^2 = 1 \times 10^{-5}$.
We discretize the control space $\mathcal{U}$ into $n_\mathcal{U} = 999$ discrete controls, $\mathcal{U} = \{ u_{min} + \frac{k (u_{max} - u_{min})}{n_\mathcal{U} - 1} \given k = 0, 1, 2, \ldots, n_\mathcal{U} - 1 \}$, using control limits $u_{min} = -1.0, u_{max} = 1.0$.
A multi-joint dynamical system has control space $\mathcal{U}^{D_u}$ and observation space $\mathbb{R}^{D_y}$ with dimensions $D_y > 1$ and $D_u > 1$, respectively. Since we couple ARX-EFE agents with single input and single output, we require $D_y = D_u$ agents to control and observe the system.
In the case of a double mass-spring-damper system, $D_y = D_u = 2$.

\subsection{Comparisons}
We compare a set of coupled ARX-EFE agents, referred to as CARX-EFE agents, with a set of uncoupled ARX-EFE agents.
CARX-EFE and uncoupled ARX-EFE agents differ in the size $M$ of the history vector $x_k$.
For each buffer type, we use a history size of $2$.
Thus, an uncoupled ARX-EFE agent has a memory size $M = 4$ ($2$ each for a history of its own observations and controls). CARX-EFE has a memory size of $M = 6$, as we additionally include a history of observations of the other agent.
Each agent has a set of parameters $(\mu_0, \Lambda_0, \alpha_0, \beta_0, \eta_0)$.
By initializing $\mu_0$ as a zero matrix and $\Lambda_0$ as an identity matrix (each of size $M$), we ensure initial conditions for optimization that give each element in $x_t$ equal importance to calculate the control objective in Eq.~\ref{eq:EFE-final}.
We further choose $\alpha_0 = 2.0$, $\beta_0 = 3.0$, and $\eta_0 = 0.001$.
The parameters of the goal priors for each agent are $(m_{1,*}, v_{1,*}) = (1.0, 1.0)$ and $(m_{2,*}, v_{2,*}) = (2.0, 1.0)$.

\subsection{Results}

\begin{figure}[htbp]
\centering
\begin{subfigure}[b]{1.0\textwidth}
    \centering
    \includegraphics[width=\textwidth]{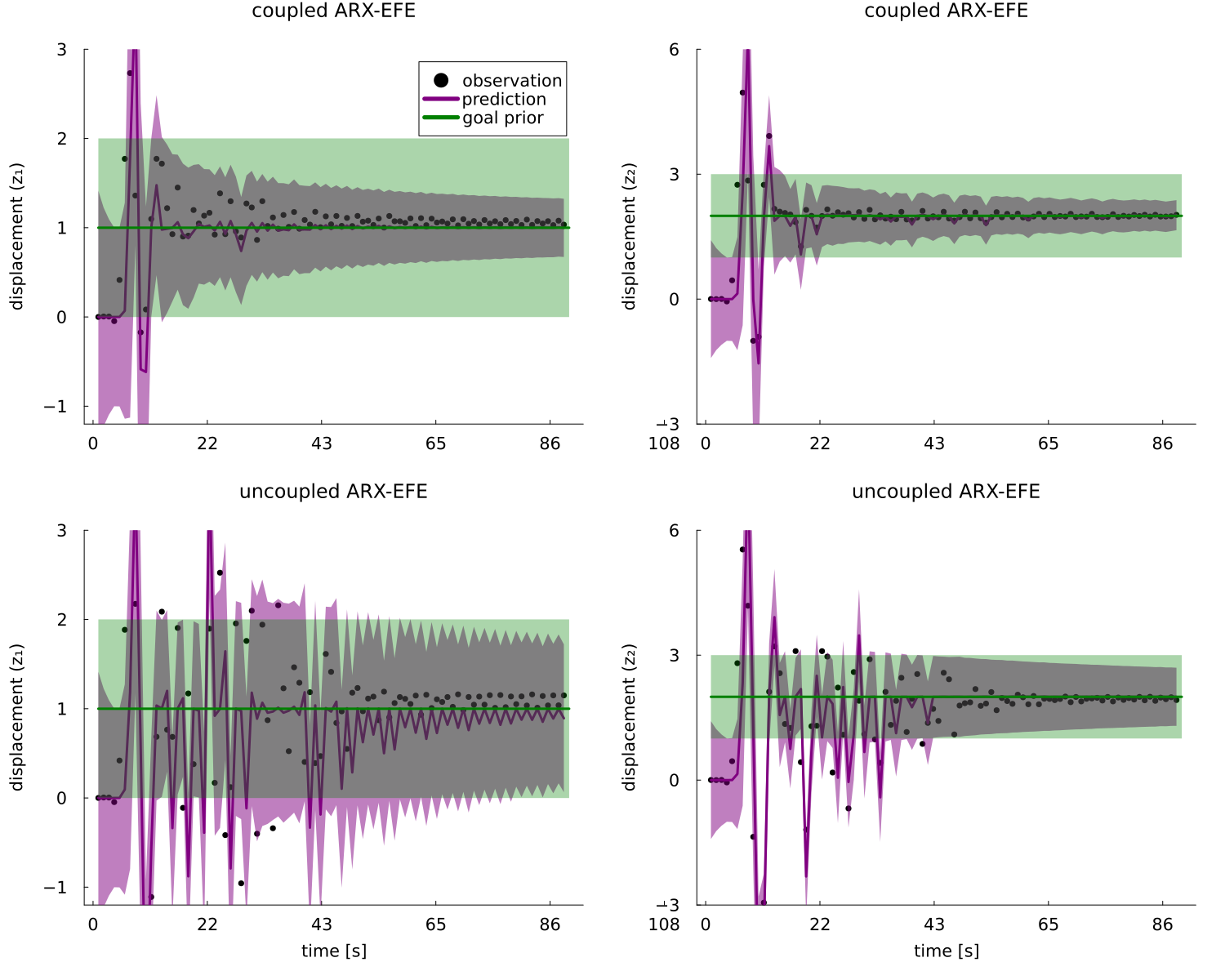}
    \caption{Observations (scatter points) and predictions of displacements of the two blocks (left = displacement $z_1$ of mass $m_1$, right = displacement $z_2$ of mass $m_2$, in purple), plotted over time. Goal prior distributions plotted in green. Both the prediction and goal prior variance are indicated by a shaded ribbon corresponding to one standard deviation. Compared to its uncoupled counterpart, CARX-EFE achieves lower prediction uncertainty, as indicated by lower prediction variance.}
    \label{fig:experiments:CARX:observations}
\end{subfigure}

\vspace{1em}

\begin{subfigure}[b]{1.0\textwidth}
    \centering
    \includegraphics[width=\textwidth]{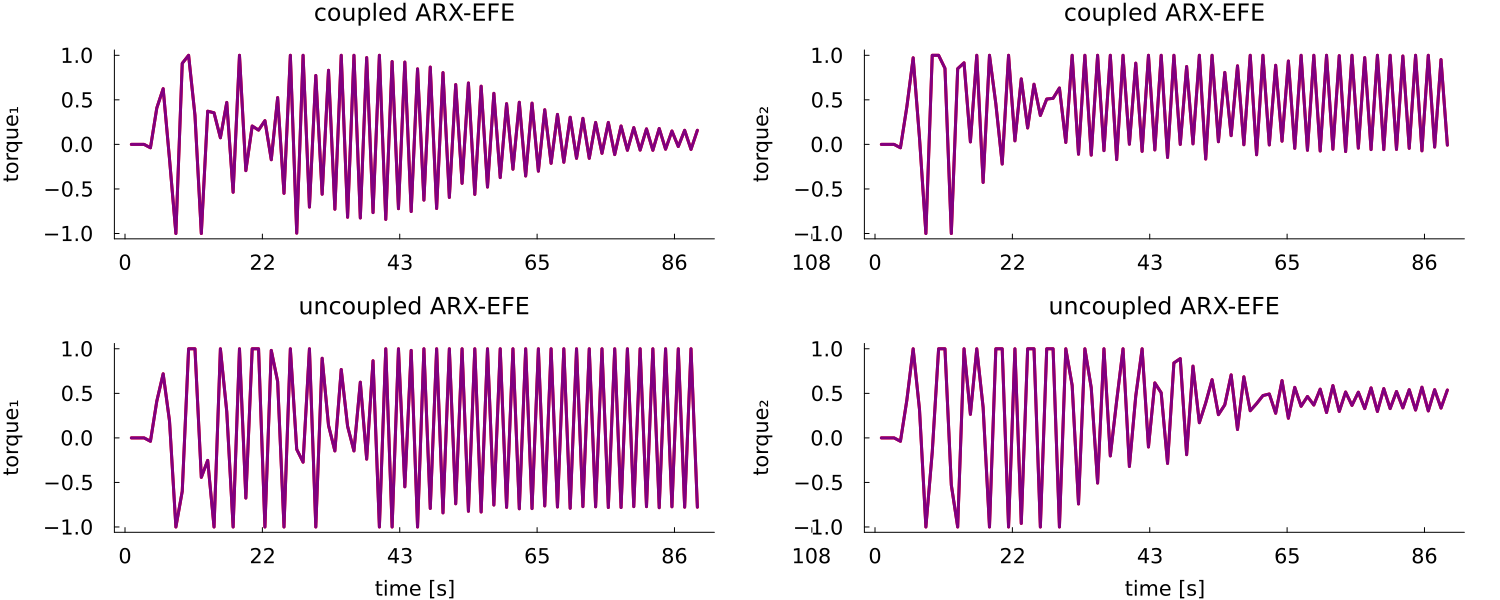}
    \caption{Controls plotted over time. All agents exhibit a short inactivity phase in the beginning, before choosing non-zero controls. The control signals for both the coupled agent controlling mass $m_1$ and the uncoupled agent controlling mass $m_2$ have an initial period of large oscillations, which gradually diminish in amplitude, eventually converging to specific values ($0.0$ for the coupled agent, $0.4$ for the uncoupled agent) with narrower oscillations.}
    \label{fig:experiments:CARX:controls}
\end{subfigure}

\caption{Comparison of predictions and controls of a set of CARX-EFE agents (top rows) versus a set of uncoupled ARX-EFE agents (bottom rows). Each column represents an agent controlling the first and second mass, respectively.}
\label{fig:experiments:CARX:predictions-controls}
\end{figure}

Figure \ref{fig:experiments:CARX:observations} shows the displacements of the two masses ($z_1$ for mass $m_1$ on the left and $z_2$ for mass $m_2$ on the right) as a function of time for the coupled agents (top row) compared to the uncoupled agents (bottom row).
The black scatter points show the observations that the system generated, while the agent's one-step ahead predictions are shown as purple lines, accompanied by ribbons indicating one standard deviation of the prediction variance. 
The goal prior, indicating the desired displacement over time, is shown in green with a ribbon reflecting one standard deviation of the goal prior variance.
The CARX-EFE agents demonstrate rapid stabilization around the goal prior, with displacements converging towards the goal prior within the first $20$ time steps.
After reaching the goal prior, oscillations around it diminish over time, resulting in a stable state where both displacements remain within a narrow range of the desired values, as indicated by the low prediction variance.
In contrast, the uncoupled ARX-EFE agents oscillate more wildly (until around time step $45$) and have more difficulty maintaining close adherence to the goal prior.
They exhibit a prolonged oscillatory phase, where oscillations are more persistent and take significantly longer to dampen.
The higher prediction variance further highlights the increased uncertainty and instability in the performance of uncoupled ARX-EFE agents compared to CARX-EFE.
The control signals (Fig.~\ref{fig:experiments:CARX:controls}) provide further insight into the observed differences in stabilization performance.
Both sets of agents start with a brief initial phase of inactivity, during which the control signals remain at zero, keeping  the system in its initial, stable state.
Following this inactivity phase, both sets of agents apply non-zero control inputs characterized by relatively large oscillations where they learn the input-output relationship before moving to the goal prior.
After reaching the goal prior, the control pulse width of one agent in each set gradually converges to specific values ($0.0$ for the coupled agent, $0.4$ for the uncoupled agent), while the other agent in the set alternates between a high and a low control value of the control space $\mathcal{U}$.
These oscillations are more narrow for the agent in control of mass $m_2$, compared to the uncoupled ARX-EFE agent controlling mass $m_1$. 

\begin{figure}[htbp]
\centering
\begin{subfigure}[b]{1.0\textwidth}
    \centering
    \includegraphics[width=\textwidth]{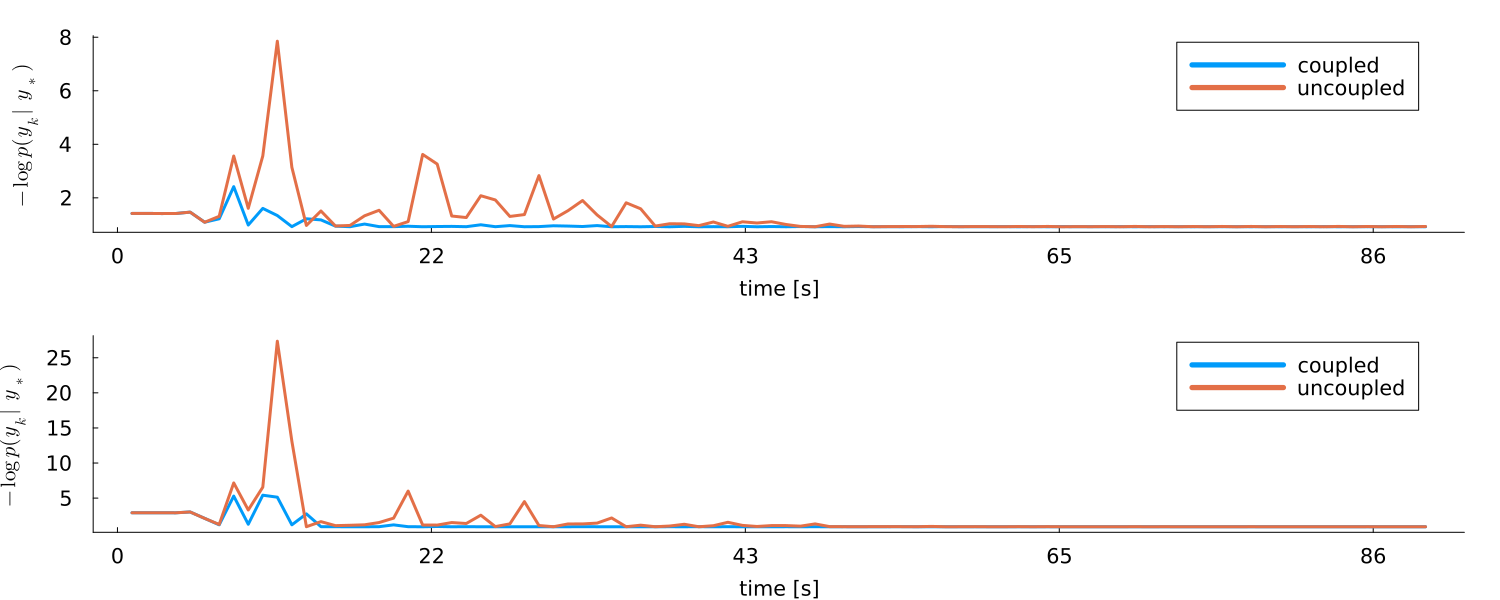}
    \caption{Goal alignment, measured by $-\log p(y_t \given y_*)$, plotted over time. Coupled agents have better overall goal alignment, with less fluctuations compared to uncoupled agents.}
    \label{fig:experiments:CNARX:performance:goal-alignment}
\end{subfigure}

\begin{subfigure}[b]{1.0\textwidth}
    \centering
    \includegraphics[width=\textwidth]{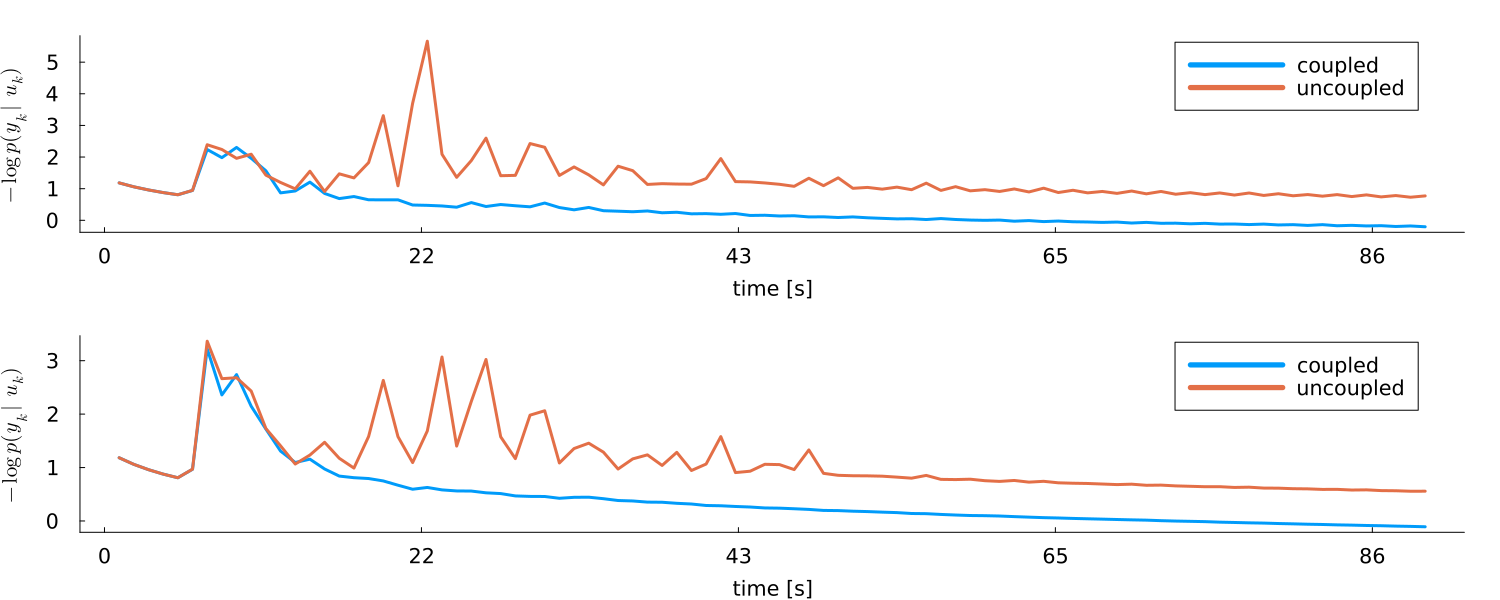}
    \caption{Prediction error (surprise), measured by the negative log-likelihood $- \log p(y_t \given u_t)$, plotted over time. CARX-EFE agents achieve better performance by minimizing surprise more effectively.}
    \label{fig:experiments:CNARX:performance:surprise}
\end{subfigure}
\caption{Comparison of model performance of a set of CARX-EFE agents versus a set of uncoupled ARX-EFE agents. Each subplot evaluates a specific aspect of performance: (a) goal alignment and (b) prediction error (surprise). Lower values indicate better performance. The top and bottom row in each subplot show the performance of agents controlling the first and second mass, respectively.}
\label{fig:experiments:CARX:performance}
\end{figure}

Figure \ref{fig:experiments:CARX:performance} compares the model performance of both agent sets over time, divided into two subplots: goal alignment (Fig.~\ref{fig:experiments:CNARX:performance:goal-alignment}) and surprise (Fig.~\ref{fig:experiments:CNARX:performance:surprise}).
Each subplot is further split into two rows, showing the performance of agents controlling the first and second mass, respectively.
Goal alignment, quantified as $- \log p(y_t \given y_*)$, measures how closely the agent's predictions align with the desired outcome (goal prior).
As illustrated in Figure \ref{fig:experiments:CNARX:performance:goal-alignment}, the CARX-EFE agents consistently achieve better goal alignment over time, compared to the uncoupled ARX-EFE agents.
Both agent sets exhibit initial peaks in the alignment error, reflecting difficulty in achieving goal alignment during the early stages of control.
For the uncoupled agents, these peaks are notably larger and more frequent, reflecting greater initial instability and less effective goal adherence.
Over time, the CARX-EFE agents maintain more stable and lower error values, suggesting a more robust alignment with the desired system state.
Prediction error, measured by $- \log p(y_t \given u_t)$, reflects the agent's ability to minimize surprise by accurately predicting system behavior based on control inputs.
Figure \ref{fig:experiments:CNARX:performance:surprise} demonstrates that the CARX-EFE agents outperform the uncoupled ARX-EFE agents by consistently achieving lower surprise values.
This suggests that CARX-EFE agents are more effective in learning the system dynamics and predicting the outcome of their actions, which in turn helps maintain better goal adherence.
In contrast, the uncoupled agents, initially struggling with higher surprise values, demonstrate less accurate predictions over time.

Overall, CARX-EFE agents exhibit superior performance by improving stabilization, lower prediction variance, and more efficient control strategies compared to their uncoupled counterparts.
These findings underscore the efficacy of the coupled approach in improving both the accuracy and stability of the control system, making CARX-EFE a more robust choice for managing complex dynamical systems.

\section{Discussion} \label{sec:discussion}
Improved ability to stabilize and lower prediction variance demonstrated by CARX-EFE suggest a significant advantage in scenarios requiring reliable convergence, such as robotic control and adaptive systems in unpredictable environments.
However, the current findings are based on a single simulation run, necessitating further validation.
Conducting Monte Carlo experiments would confirm the robustness of CARX-EFE's advantages across varied conditions.
Future work should also evaluate the CARX-EFE agents on nonlinear and underactuated systems, like a double pendulum or acrobot, to assess their ability to generalize.
Additionally, benchmarking against other control methods could provide insights into the relative strength of CARX-EFE agents.
The current implementation of CARX-EFE agents relies on a one-step ahead prediction, making their performance sensitive to the system update step size ($\Delta t$).
Addressing this limitation by extending the prediction capability could reduce the dependence on these parameters, and possibly improve the efficiency of the coupled approach.

\section{Conclusion}
We investigated the control of a multi-joint mechanical system by coupling multiple autoregressive active inference agents that minimize expected free energy.
We evaluate the effect of sharing data buffers (i.e., memories) in the autoregressive models of the agents.
Our experiments demonstrate that coupling significantly improves the agent's ability to achieve both better goal alignment and lower prediction error.
CARX-EFE agents consistently outperformed their uncoupled counterparts, showing lower prediction uncertainty with higher prediction accuracy (lower surprise), and greater long-term stability around the goal prior.
It is important to note that the agent is limited to one-step ahead predictions.
Future research should focus on extending the horizon of the agents, and improving the optimization procedure in MAP estimation.

\begin{credits}
\subsubsection{\ackname} 
The authors gratefully acknowledge support by the Eindhoven Artificial Intelligence Systems Institute and the Ministry of Education, Culture and Science of the Government of the Netherlands.

\subsubsection{\discintname}
The authors have no competing interests to declare that are
relevant to the content of this article. 
\end{credits}

\bibliographystyle{splncs04}
\bibliography{camready}

\end{document}